\documentclass[10pt,twocolumn,letterpaper]{article}

\usepackage{iccv}              

\usepackage [accsupp]{axessibility}
\usepackage{multirow}
\usepackage{makecell}
\usepackage{bm}
\definecolor{iccvblue}{rgb}{0.21,0.49,0.74}
\usepackage[pagebackref,breaklinks,colorlinks,allcolors=iccvblue]{hyperref}

\def\paperID{6239} 
\def\confName{ICCV}
\def\confYear{2025}

\title{Accelerating Diffusion Transformer via Gradient-Optimized Cache}

\author{
 \textbf{Junxiang Qiu\textsuperscript{1}},
 \textbf{Lin Liu\textsuperscript{1}},
 \textbf{Shuo Wang{\textsuperscript{1}\thanks{Shuo Wang is the corresponding author}}},
 \textbf{Jinda Lu\textsuperscript{1}},
 \textbf{Kezhou Chen\textsuperscript{1}},
 \textbf{Yanbin Hao\textsuperscript{2}}
\\
 \small{\textsuperscript{1}University of Science and Technology of China;
 \textsuperscript{2}Hefei University of Technology.}
\\
 \small{
   \{qiujx, liulin0725, lujd, chenkezhou\}@mail.ustc.edu.cn, 
   shuowang.edu@gmail.com, haoyanbin@hotmail.com
 }
}

\begin{document}
\maketitle
\begin{abstract}
Feature caching has emerged as an effective strategy to accelerate diffusion transformer (DiT) sampling through temporal feature reuse. It is a challenging problem since (1) Progressive error accumulation from cached blocks significantly degrades generation quality, particularly when over 50\% of blocks are cached; (2) Current error compensation approaches neglect dynamic perturbation patterns during the caching process, leading to suboptimal error correction. To solve these problems, we propose the Gradient-Optimized Cache (GOC) with two key innovations: 
(1) Cached Gradient Propagation: A gradient queue dynamically computes the gradient differences between cached and recomputed features. These gradients are weighted and propagated to subsequent steps, directly compensating for the approximation errors introduced by caching. (2) Inflection-Aware Optimization: Through statistical analysis of feature variation patterns, we identify critical inflection points where the denoising trajectory changes direction. By aligning gradient updates with these detected phases, we prevent conflicting gradient directions during error correction.
Extensive evaluations on ImageNet demonstrate GOC's superior trade-off between efficiency and quality. With 50\% cached blocks, GOC achieves IS 216.28 (26.3\%↑) and FID 3.907 (43\%↓) compared to baseline DiT, while maintaining identical computational costs. These improvements persist across various cache ratios, demonstrating robust adaptability to different acceleration requirements.
Code is available at \url{https://github.com/qiujx0520/GOC_ICCV2025.git}.
\end{abstract}    
\section{Introduction}
\label{sec:intro}
Diffusion Transformers (DiT) \cite{peebles2023scalable,gao2023masked,yang2024cogvideox,lu2023semantic,qiu2025accelerating} have shown a powerful ability in content generation in synthesizing multimodal content spanning textual\cite{lu2025damo,zhu2024selective}, visual\cite{wang2025precise,zhu2024boosting,zhu2024enhancing}, and temporal\cite{qiu2025multimodal,wang2018connectionist,guo2019dense} domains. It is widely applied in numerous fields, including intelligent creation assistance, information processing, and digital entertainment content generation. However, it is time-consuming to sample during their forward and reverse processes \cite{ho2020denoising,selvaraju2024fora}. This significantly hinders the technology's rapid deployment and flexible application in the real world. Therefore, accelerating the content generation process and improving its inference efficiency have become an important issue \cite{ma2024deepcache}.

\begin{figure}[t]
  \centering
   \includegraphics[width=1\linewidth]{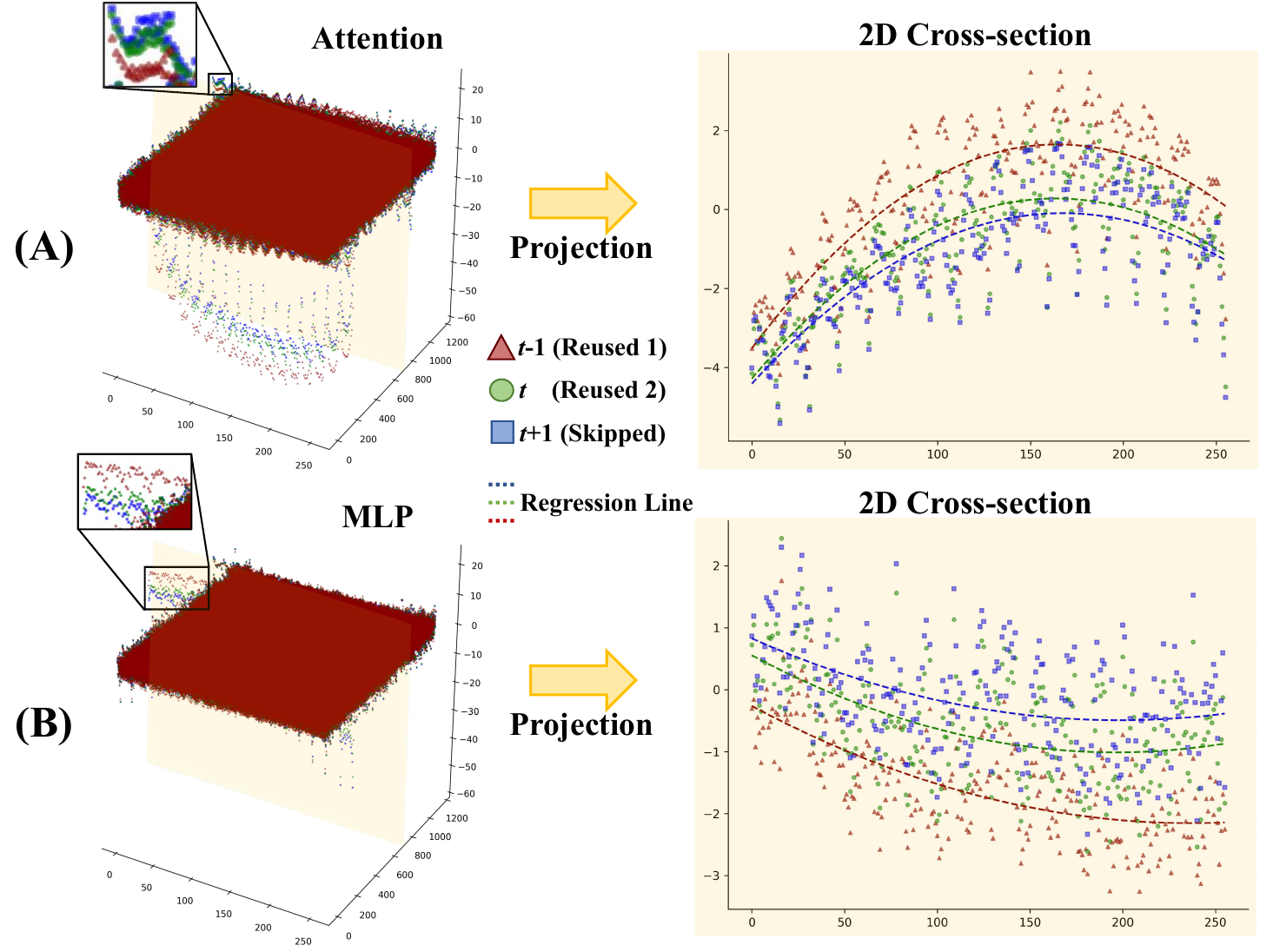}
   \caption{Outputs of the two most recently reused steps $t-1$, $t$ (red triangles and green circles) and the step $t+1$ (blue squares) that will be skipped in the DiT sample process for the (A) attention layer and the (B) MLP layer. Meanwhile, to better observe the direction of these points, we also provide the fitting curves of these points (in red, green, and blue dotted lines).}
   \label{fig:blockview}
\end{figure}

Recent studies usually use pruning \cite{liu2021group} or caching \cite{selvaraju2024fora} strategies to accelerate the generation process.
In contrast to pruning techniques that achieve acceleration through structural simplification by removing redundant parameters or deactivating non-critical neurons, caching-based approaches leverage the temporal coherence in sequential generation processes to reuse intermediate computational states across multiple sampling steps. This paradigm shift preserves the model's original expressive capacity and style characteristics while significantly enhancing sampling efficiency through computational reuse.
However, current caching implementations exhibit a critical limitation: The prevalent practice of direct state reuse in consecutive steps lacks rigorous error analysis of cached computations \cite{selvaraju2024fora,ma2024deepcache,wimbauer2024cache}. Specifically, the recursive application of cached features introduces error propagation in multi-step generation processes, where approximation errors accumulate through successive caching operations.

Smoothcache\cite{liu2024smoothcache} uses numerical gradients at different time steps in the sampling process to locate modules suitable for caching, without discussing how to reduce the errors that have already occurred. Knowledge editing\cite{meng2022locating} affects generated content by introducing perturbations. These works provide us with inspiration for reducing caching errors. We believe that these caching errors can be offset by incorporating these gradients into the caching process. 

To intuitively illustrate this, we visualized the outputs of the Attention and MLP layers during the DiT computation process in Figure \ref{fig:blockview}(A) and (B), respectively, where the red triangles, green circles, and blue squares respectively represent: the outputs from the two most recently reused computational steps, and the output of the step to be skipped in subsequent calculations. 
Previous caching methods typically employ the green circles (recently reused outputs) to directly substitute the blue squares (computations to be skipped). However, we identify a persistent yet systematic positional deviation between these geometrically similar patterns. Our visualization analysis reveals that incorporating the temporal dimension of the red triangles (historical reused states) exposes a critical directional relationship: the compensation error manifests as a vector space displacement where adjusting the green circle's position along the inverse gradient direction of the red triangle can effectively minimize its deviation from the target blue square. Thus, we leverage the gradient compensation mechanism to strategically integrate gradients from the two most recent reused computations to adaptively adjust cached content, thereby addressing the positional drift in next-step cache generation.

Specifically, we first propose a gradient caching method to reduce caching errors. Concurrently, we leverage statistical information from the model to guide gradient caching implementation, avoiding inverse gradient optimization pitfalls while ensuring high-quality generation. To achieve this, we integrate a queue mechanism into the caching process. By calculating gradients between the two most recently cached contents and incorporating them into subsequent cached content, we align reused features more closely with skipped features, thereby minimizing caching errors.
Additionally, we observe that a subset of skipped features exhibit gradient direction misalignment with cached gradients. Applying gradient optimization to these blocks introduces noise. Notably, perturbations introduced in early sampling steps allow partial error correction through subsequent mappings, resulting in limited sampling side effects. In contrast, late-step perturbations propagate errors directly to the final generated image.

To further enhance Gradient Cache's error elimination capability, we calculate inverse-gradient block proportions per step using model statistics. Combining these proportions with step positions in the sampling process, we devise a gradient cache optimization decision strategy. This approach prevents major error introduction while confining residual minor errors to later sampling stages, thereby maximizing gradient optimization benefits. We call the entire cache calculation process Gradient-Optimized Cache (GOC).
In summary, the contributions are threefold:
\begin{itemize}
\item We design a new basic model caching strategy that leverages the unique gradients in the caching process to reduce errors in future caching.
\item We propose a method to determine when gradient cache optimization needs to be applied, aiming to reduce the errors associated with gradient cache optimization.
\item Our method reduces caching errors and optimizes the generation quality. It can increase the upper limit of the image generation speed without sacrificing quality or introducing additional computational costs. 
\end{itemize}

\section{Related Work}
\label{sec:relatedwork}
In this section, we first review existing diffusion acceleration methods and then outline the differences between our proposed GOC and these related methods.

\subsection{Traditional Sampling Acceleration Method}
Common approaches for accelerating sampling in diffusion models fall into three categories: pruning, quantization, and efficient sampling methods.
\textbf{Pruning} reduces model complexity by removing less critical components while maintaining performance. Methods are broadly categorized into unstructured pruning \cite{dong2017learning,lee2019signal}, which masks individual parameters, and structured pruning \cite{liu2021group}, which eliminates larger structures like layers or filters. DiP-GO \cite{zhu2025dip} introduces a plugin pruner that optimizes pruning constraints to maximize synthesis capability. DaTo \cite{zhang2024token} dynamically prunes tokens with low activation variance, retaining only high-variance tokens for self-attention computation, thereby enhancing temporal feature dynamics. 
\textbf{Quantization} compresses models by representing weights and activations in lower-bit formats. Key strategies include Quantization-Aware Training (QAT) \cite{bhalgat2020lsq}, which embeds quantization into training, and Post-Training Quantization (PTQ) \cite{li2021brecq,nagel2020adaptive}, which directly quantizes pre-trained models without retraining. For diffusion models, Q-Diffusion \cite{li2023qdiffusion} improves calibration through time step-aware data sampling and introduces a specialized noise-prediction network quantizer. PTQ4DiT \cite{wu2024ptq4dit} achieves 8-bit (W8A8) quantization for Diffusion Transformers (DiTs) with minimal quality loss and pioneers 4-bit weight quantization (W4A8).
\textbf{Efficient Sampling} seeks to minimize computational overhead while preserving generation quality in diffusion models. This is primarily through two paradigms: retraining-based optimization and sampling-algorithm enhancement. Retraining approaches like knowledge distillation \cite{salimans2022progressive,chadebec2024flash} modify model architectures to enable fewer-step generation, albeit at the cost of additional training resources. Conversely, training-free methods focus on refining sampling dynamics. Notably, DDIM \cite{song2020denoising} accelerates inference via non-Markovian deterministic trajectories, while DPM-Solver \cite{lu2022dpm} employs high-order differential equation solvers to reduce step counts theoretically. Meanwhile, consistency models \cite{song2023consistency} enable single-step sampling through learned transition mappings. Parallel frameworks like DSNO \cite{zheng2023fast} and ParaDiGMS \cite{shih2024parallel} exploit temporal dependencies between denoising steps for hardware-accelerated throughput.

\subsection{Model Caching}
In addition to the aforementioned methods, model caching provides a low-cost, efficient, and versatile approach for accelerating diffusion generation. Common caching strategies are primarily divided into two categories: rule-based \cite{selvaraju2024fora,ma2024deepcache,wimbauer2024cache,qiu2025accelerating} methods, which reuse or skip specific steps/blocks by analyzing sampling-induced feature variations, and training-based \cite{ma2024learning} methods, where models learn to skip non-critical modules through training. These strategies are widely integrated into DiT architectures due to their strong learning capabilities, and their effectiveness is further facilitated by the unchanged data dimensionality during sampling. Initially, U-Net-based methods like DeepCache \cite{ma2024deepcache} and Faster Diffusion \cite{li2023faster} achieved low-loss computation skipping through feature reuse, while Cache-Me-if-You-Can \cite{wimbauer2024cache} reduced caching errors via teacher-student mimicry. However, such techniques are challenging to adapt directly to DiT. To extend caching to DiT-based models, Fora \cite{selvaraju2024fora} stores and reuses attention/MLP layer outputs across denoising steps, $\Delta$-DiT \cite{chen2024delta} accelerates specific blocks via dedicated caching. Additionally, Learning-to-Cache \cite{ma2024learning} employs a trainable router to dynamically skip layers, achieving higher acceleration ratios but incurring significant computational costs.

\noindent\textbf{Differences:} 
Based on the analysis of related work, our approach falls under the branch of model caching that focuses on optimizing cache errors. The method most closely related to ours is Fora \cite{selvaraju2024fora}. The key differences between our method and the aforementioned methods are threefold:
Firstly, the existing caching methods primarily aim to accelerate sampling by locating and skipping weakly correlated layers. In contrast, our method builds upon their caching mechanisms and leverages gradients to reduce cache errors.
Secondly, in GOC, we conduct statistical analysis on the output of each block in the model. Combined with step positions, this approach reduces gradient errors and maximizes the benefits brought by gradient caching.
Thirdly, our method can be applied to various existing rule-based and training-based methods, and it can enhance the performance of these methods.
\section{Method}

\begin{figure*}[t]
  \centering
   \includegraphics[width=1\linewidth]{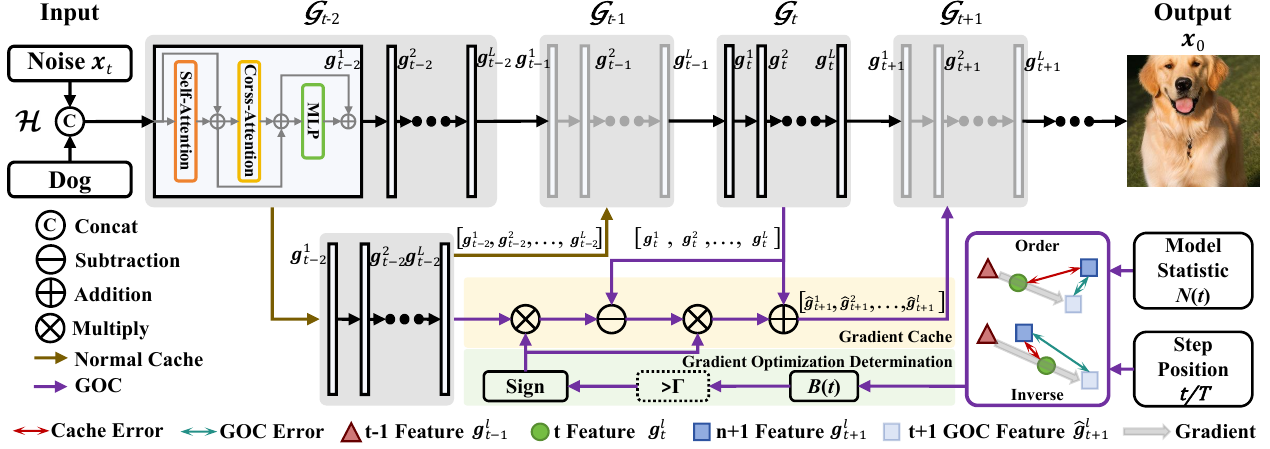}
   \caption{Pipeline of GOC. The input of GOC includes four parts: features $\bm{g}_{t-1}^l$, $\bm{g}_{t}^l$ from the most recent two Step Caches, model statistical information $\bm{N}(t)$, and step position $t/T$. Gradient Cache: the most recent two features are used to calculate the gradient, which is then multiplied by a coefficient and added back to the most recent feature, replacing the feature of the next step that will be skipped. Gradient Optimization Determination: By calculating the weighted sum of the model's statistical information and the step position, and comparing it with a pre-set threshold, it determines whether to use the gradient values computed by GC in the next step.}
   \label{fig:pipeline}
\end{figure*}

In this section, we first briefly revisit the preliminaries of the diffusion method. Second, we illustrate our Gradient-Optimized Cache (GOC) in three parts: 
(1) Gradient Cache (GC), (2) Model Information Extraction and Statistics, and (3) Gradient Optimization Determination (GOD).

\subsection{Preliminaries}

\textbf{Diffusion Models}. Diffusion models \cite{ho2020denoising,song2020denoising} learn denoising methods to restore random noise $\bm{x_t}$ back to the original image $\bm{x_0}$. To achieve this step-by-step, the model needs to learn the reverse process of adding noise. Using Markov chains $\bm{\mathcal{N}}$, the reverse process $\bm{p_{\theta}(x_{t-1} \mid x_{t})}$ can be modeled as follows:
\begin{equation}
\bm{\mathcal{N}} \left( \bm{x}_{t-1}; \frac{1}{\sqrt{\alpha_t}} \left( \bm{x}_t-\frac{1-\alpha_t}{\sqrt{1-\bar{\alpha}_t}} \bm{\epsilon}_{\bm{\theta}}(\bm{x}_t, t) \right) \right),
\label{eq:1}
\end{equation}
where $ t $  represents the denoising step, $ \beta_t $ denotes the noise variance schedule, $ \alpha_t = 1 - \beta_t $, $ \bar{\alpha}_t = \prod_{i=1}^{T} \alpha_i $, and $ \bm{T} $ represents the total number of denoising steps.
$\bm{\epsilon}_{\bm{\theta}}$ is a deep network parameterized by $\bm{\theta}$. It takes $\bm{x}_t$ as input and outputs the prediction of the noise required for the denoising process. The aforementioned repeated inference is the primary source of computational cost in diffusion models.

\noindent\textbf{Diffusion Transformer}. Diffusion Transformer (DiT) \cite{peebles2023scalable} typically consists of stacked self-attention layers $\bm{S}$, cross-attention layers $\bm{C}$, and multilayer perceptrons (MLP) $\bm{M}$, all with identical dimensionality. It can be described by the following formulation:
\begin{equation}
\begin{split}
\bm{\mathcal{H}}&=\bm{\mathcal{G}}_1\circ\bm{\mathcal{G}}_2\circ\cdots\circ\bm{\mathcal{G}}_l\circ\cdots\circ\bm{\mathcal{G}}_T,\quad\text{where}\\
\bm{\mathcal{G}}_t&=\bm{g}_t^1\circ \bm{g}_t^2\circ\cdots\circ\bm{g}_t^l \circ\cdots\circ\bm{g}_t^L,\quad\text{where}\\
\bm{g}_t^l&={\bm{S}}_t^l\circ{\bm{C}}_t^l\circ{\bm{M}}_t^l,
\end{split}
\label{eq:2}
\end{equation}
where $\bm  {\mathcal{H}}$ represents the entire DiT process, $\bm{\mathcal{G}}_t$ denotes the process at the $t$-th step, and $L$ represents the depth of the DiT model at each step. $\circ$ is the elements-wise product. $\bm{g}_t^l$ refers to the $l$-th module in the DiT architecture at the $t$-th step, while $\bm{S}_t^l$, $\bm{C}_t^l$ and $\bm{M}_t^l$ represent the self-attention layer, cross-attention layer, and MLP in a single DiT block, respectively.
Subsequently, the general computation between  $\bm{S}_t^l$, $\bm{C}_t^l$, $\bm{M}_t^l$ and $\bm{x}_t^l$ can be written as:
\begin{equation}
\begin{split}
\bm{S}_t^l&=\bm{x}_t^l+\text{AdaLN}\circ \bm{s}_t^l(\bm{x}_t^l),\\
\bm{C}_t^l&=\bm{S}_t^l+\text{AdaLN}\circ \bm{c}_t^l(\bm{S}_t^l),\\
\bm{x}_{t+1}^l=\bm{M}_t^l&=\bm{C}_t^l+\text{AdaLN}\circ \bm{m}_t^l(\bm{C}_t^l),
\end{split}
\label{eq:f}
\end{equation}
where $\bm{x}_t^l$ represents the residual connection, and $\bm{s}_t^l$, $\bm{c}_t^l$, $\bm{m}_t^l$ denote the computation of the self-attention layer, cross-attention layer, and MLP layer, respectively. AdaLN \cite{guo2022adaln}
is applied after the computations of $\bm{s}_t^l(\bm{x}_t^l)$, $\bm{c}_t^l(\bm{x}_t^l)$ and $\bm{m}_t^l(\bm{x}_t^l)$, which helps stabilize the training process and enhances the overall performance of the DiT model in handling complex data patterns.

\noindent\textbf{Feature Caching and Reuse}.  
Feature caching \cite{ma2024deepcache} aims to reuse time-consuming computation results from past steps to skip and replace future step computations. 
In our method, if the feature is fully computed at step \( t \), then the computation required at step \( t+1 \) will be skipped and directly replaced by the computation from step \( t \). 
Taking the \( l \)-th block as an example, the caching process can be expressed as \(\bm{U}[l] := [\bm{s}_t^l(\bm{x}_t^l), \bm{c}_t^l(\bm{x}_t^l), \bm{m}_t^l(\bm{x}_t^l)]\), where “\(:=\)” denotes the assignment operation. \(\bm{U}[l]\) caches the computation outputs of the attention layer and the MLP layer from the \( l \)-th block at step \( t \). Subsequently, the computation at step \( t+1 \) is skipped by reusing the cached features, which can be represented as \([\bm{s}_{t+1}^l(\bm{x}_{t+1}^l), \bm{c}_{t+1}^l(\bm{x}_{t+1}^l), \bm{m}_{t+1}^l(\bm{x}_{t+1}^l)] := \bm{U}[l]\).

\subsection{Gradient-Optimized Cache}


The overview of our framework is depicted in Figure \ref{fig:pipeline}. 
To minimize the error introduced by caching, we compute the gradients of the features from the two most recent steps, $\bm{g}_{t-1}^l$ and $\bm{g}_{t}^l$, multiply them by a coefficient $\theta$, and add them to the features of $\bm{g}_{t}^l$. These modified features are then used to replace the features of the next skipped step $t+1$. Additionally, we determine whether to apply normal caching or GOC to a skipped step by comprehensively analyzing the model's statistical information and the position of the step. Below, we introduce the three steps involved in GOC.

\noindent\textbf{Gradient Cache (GC)}.
Storing features in two steps is necessary to obtain gradients and enable gradient caching. Therefore, we introduce a queue to simultaneously store the features from the two most recent steps, $\bm{g}_{t-1}^l$ and $\bm{g}_{t}^l$. In the early stages of the sampling process when gradients have not yet formed, as indicated by the brown arrow in Figure \ref{fig:pipeline}, we employ normal caching as:
\begin{equation}
\bm{g}_{1}^l := \bm{g}_{0}^l.
\label{eq:gc}
\end{equation} 
This involves reusing the results of $\bm{g}_{0}^l$ and skipping the computation of $\bm{g}_{1}^l$.
When $ t > 3 $, we use the cached features from the two steps preceding the skipped step to compute the final reused feature. The computation method of Gradient Cache (GC) is as follows:
\begin{equation}
\bm{g}_{t+1}^l := \hat{\bm{g}}_{t+1}^l := \bm{g}_{t}^l+\eta*(\bm{g}_{t}^l-\bm{g}_{t-1}^l),
\label{eq:gc}
\end{equation}
where $\hat{\bm{g}}_{t+1}^l$ is the feature computed from $\bm{g}_{t+1}^l$ that will be reused, and $\eta$ is a positive parameter used to adjust the magnitude of the gradient.

\noindent\textbf{Model Information Extraction and Statistics}.
However, excessive application of GC introduces additional gradient noise during error mitigation, ultimately degrading image generation quality. The fundamental source of this gradient noise stems from directional conflicts in gradient vectors - specifically, the expected input vector $\bm{g}_{t+1}^l$ (represented by the blue square in Figure \ref{fig:pipeline}) deviates in the opposite direction from the gradient path connecting $\bm{g}_{t-1}^l$ (red triangle) to $\bm{g}_{t}^l$ (green circle). As illustrated in Figure \ref{fig:pipeline}, the computed GC feature $\hat{\bm{g}}_{t+1}^l$ (light blue square) exhibits a displacement from the ideal $\bm{g}_{t+1}^l$ position, quantified by two distinct error metrics: the mint green arrow represents the GC approximation error, while the red arrow indicates the inherent caching error in normal operations. This geometric relationship reveals that GC achieves optimal caching fidelity when gradient directions remain consistent, but introduces progressively larger approximation errors as directional conflicts between successive gradients intensify.

To avoid introducing errors, we need to analyze the data $\bm{g}_{t}^l$ for each block of the DiT model to identify steps where inverse gradients are likely to occur. We achieve this by repeatedly performing the sampling process with different prompts and recording the outputs of the self-attention layers, cross-attention layers (if present), and MLP layers, denoted as $\bm{f}_{\mathrm{sAttn}}(\bm{x})$, $\bm{f}_{\mathrm{cAttn}}(\bm{x})$, and $\bm{f}_{\mathrm{MLP}}(\bm{x})$. We collectively refer to these matrices as $\bm{F}_t^l(k)\in \mathbb{R}^{m\times n}$, where $ k $ represents different prompts, and $ m \times n $ denotes the dimensionality of the feature.
Next, we calculate the average $\bm{A}_t^l$ for each block's $\bm{F}_t^l(k)$ as follows:
\begin{equation}
\bm{A}_t^l = \frac{1}{K} \sum_{k=1}^{K} \bm{F}_t^l(k),
\label{eq:MIES1}
\end{equation}
where $ K $ represents the total number of prompts. If there exists $\eta$, it can satisfy:
\begin{equation}
\bm{J}(\bm{A}_{t+1}^l-\bm{A}_{t}^l) > \bm{J}(\bm{A}_{t+1}^l - (\bm{A}_{t}^l+\eta*(\bm{A}_{t}^l-\bm{A}_{t-1}^l))),
\label{eq:MIES2}
\end{equation}
and we believe that $\bm{g}_{t-1}^l$, $\bm{g}_{t}^l$, and $\bm{g}_{t+1}^l$ form a positive gradient if they align in the expected direction; otherwise, they form an inverse gradient. $\bm{J}$ is defined as the sum of the absolute values of each element $s$ in the $m \times n$ matrix, where a larger result indicates a greater error. 
Finally, we record the number of blocks $\mathbf{N}(t)$ exhibiting inverse gradients at each step $t$. Steps with a higher count of such blocks are more likely to introduce gradient errors.

\noindent\textbf{Gradient Optimization Determination (GOD)}.
Determining whether a step will introduce gradient errors is not sufficient by merely calculating inverse gradients. Previous studies \cite{meng2022locating,meng2022memit} have shown that perturbations introduced closer to the end of the sampling process have a more significant impact on the generated results. From another perspective, even if errors are introduced in the early stages of the sampling process, they can be mapped to correct patterns through subsequent steps. However, errors introduced in the later stages of the sampling process are difficult to eliminate and may lead to artifacts.
Therefore, we comprehensively consider the number of inverse gradients and the position of the step to measure the negative impact $\bm{B}(t)$ of introducing gradient errors at a specific step $t$:
\begin{equation}
\bm{B}(t) = \gamma*(1-\frac{t}{T})+(1-\gamma)*\bm{N}(t),
\label{eq:MIES2}
\end{equation}
where $\gamma$ is a parameter used to adjust the balance between the step position and $\bm{N}(t)$.
We set a threshold $\bm{\Gamma}$, and only when $\bm{B}(t) < \bm{\Gamma}$ do we perform GC at step $t$.


\section{Experiment}

\begin{figure*}[t]
  \centering
   \includegraphics[width=1\linewidth]{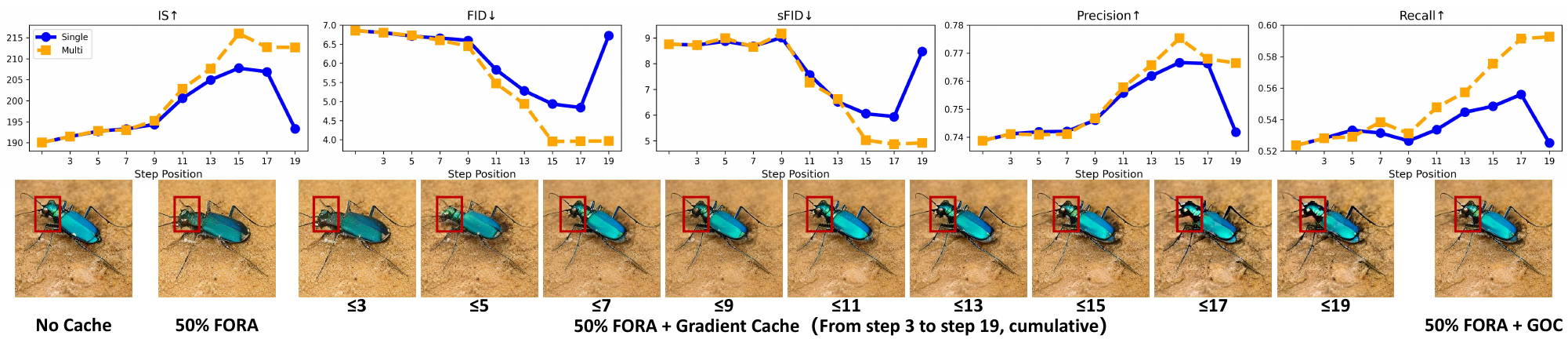}
   \caption{The line chart above shows the metrics generated by single-step GC (solid blue line) and cumulative GC (dashed orange line). And the beetle generated images below include those from No Cache, 50\% rule-based caching (FORA\cite{selvaraju2024fora}), 50\% rule-based caching (FORA) with varying strengths of GC, and 50\% FORA + GOC.}
   \label{fig:RQ1}
\end{figure*}

\begin{table*}[t]
    \centering
    \begin{tabular}{c|c|ccccc}
        \toprule
        \textbf{Cache Strategy} & \textbf{Caching Level} & \textbf{IS↑} & \textbf{FID↓} & \textbf{sFID↓} & \textbf{Precision↑} & \textbf{Recall↑}  \\ 
        \midrule
        No Cache\cite{peebles2023scalable} & 0\% & 223.490 & 3.484 & 4.892 & 0.788 & 0.571  \\ 
        \midrule
        FORA\cite{selvaraju2024fora} & \multirow{3}{*}{50\%} & 190.056 & 6.857 & 8.757 & 0.739 & 0.524\\ 
        \cline{1-1} \cline{3-7} 
        FORA+GC &  & 212.691 & 3.964 & \textbf{4.923} & 0.766 & \textbf{0.593}\\ 
        \cline{1-1} \cline{3-7} 
        FORA+GOC &  & \textbf{216.280} & \textbf{3.907} & 4.972 & \textbf{0.775} & 0.574\\ 
        \bottomrule
    \end{tabular}
    \caption{Metrics of the generated images under No Cache, as well as FORA, FORA+GC, and FORA+GOC with 50\% caching level.}
    \label{tab:RQ2}
\end{table*}

In this section, we evaluate the proposed GOC method and compare it with rule-based and training-based caching methods. We also use ablation experiments to verify the effectiveness of the method proposed in this paper. We aim to address the following research questions (\textbf{RQ}):

\noindent\textbf{RQ1}: Is Gradient Caching (GC) effective?

\noindent\textbf{RQ2}: Is Gradient Optimization Determination (GOD) effective?

\noindent\textbf{RQ3}: What hyperparameter should be selected?

\noindent\textbf{RQ4}: What are the advantages and general applicability of GOC?

\subsection{Experiment Settings}
\noindent\textbf{Datasets}. To comprehensively evaluate the performance of diffusion models, we conduct experiments on two benchmark datasets: ImageNet \cite{deng2009imagenet} (1,000 classes) for class-conditional generation and MS-COCO \cite{lin2014microsoft} (30,000 text prompts) for text-to-image synthesis. Unless explicitly stated otherwise, the ablation experiments default to using the ImageNet dataset.

\noindent\textbf{Model Configuration}. We used two models to verify the effectiveness of our method: DiT \cite{ma2024deepcache} and Pixart \cite{chen2023pixart}. For the DiT architecture, we adopt DDIM sampler \cite{song2020denoising} with 20 denoising steps throughout all experiments, while Pixart models use the DPM-Solver \cite{lu2022dpm} under the same 20-steps. To ensure fair comparison across different caching strategies, we generate 50,000 samples for DiT and 30,000 samples for Pixart respectively, matching each model's native output characteristics. The experiments are conducted on an NVIDIA A40 GPU.

\noindent\textbf{Evaluation Metrics}. We employ a comprehensive set of quantitative metrics: such as Inception Score(IS↑), Fréchet Inception Distance(FID↓), Sliced Fréchet Inception Distance(sFID↓), Precision(↑), and Recall(↑). All metrics are computed at 256×256 resolution using randomly generated samples to ensure statistical significance.

In our experiment, each cache is reused at most once. For convenience, we provide the abbreviations of Gradient Cache (GC), Gradient Optimization Determination (GOD), and Gradient-Optimized Cache (GOC). In addition, the data of L2C is completely reproduced using the open-source code provided by the authors of Learning-to-Cache\cite{ma2024learning}. And the data of FORA is implemented by summarizing the experience from previous work \cite{selvaraju2024fora}, where 25\% and 50\% of the blocks are cached.

\subsection{Gradient Cache}
\textbf{Effect of GC (RQ1)}.
To evaluate the performance characteristics of Gradient Caching (GC), Figure \ref{fig:RQ1} presents three comparative elements: single-step GC metrics (blue trajectory), multi-step cumulative GC metrics (orange trajectory), and corresponding beetle image generations. The blue trajectory demonstrates consistent metric improvement across all optimization steps compared to the baseline initialization (leftmost data point). The orange trajectory reveals progressive metric enhancement through successive GC layer integration until step 17, where we observe severe degradation in both Inception Score (IS) and precision metrics, indicative of accumulated gradient approximation errors. By comparing the beetle images generated with 50\% cache + Gradient Cache, 50\% Cache, and No Cache, we can find that 50\% cache + GC exhibit progressive quality improvements through step 15, progressively approximating the quality of the No Cache configuration. However, post-step 17 generations display pronounced artifacts in the cephalic region and thoracic segments (highlighted in red boxes). Therefore, an appropriate level of GC can overcome most errors introduced by Normal Caching. However, excessive GC introduces gradient errors and reduces image quality, making it necessary to select for GC application.

\begin{figure}[t]
  \centering
   \includegraphics[width=0.9\linewidth]{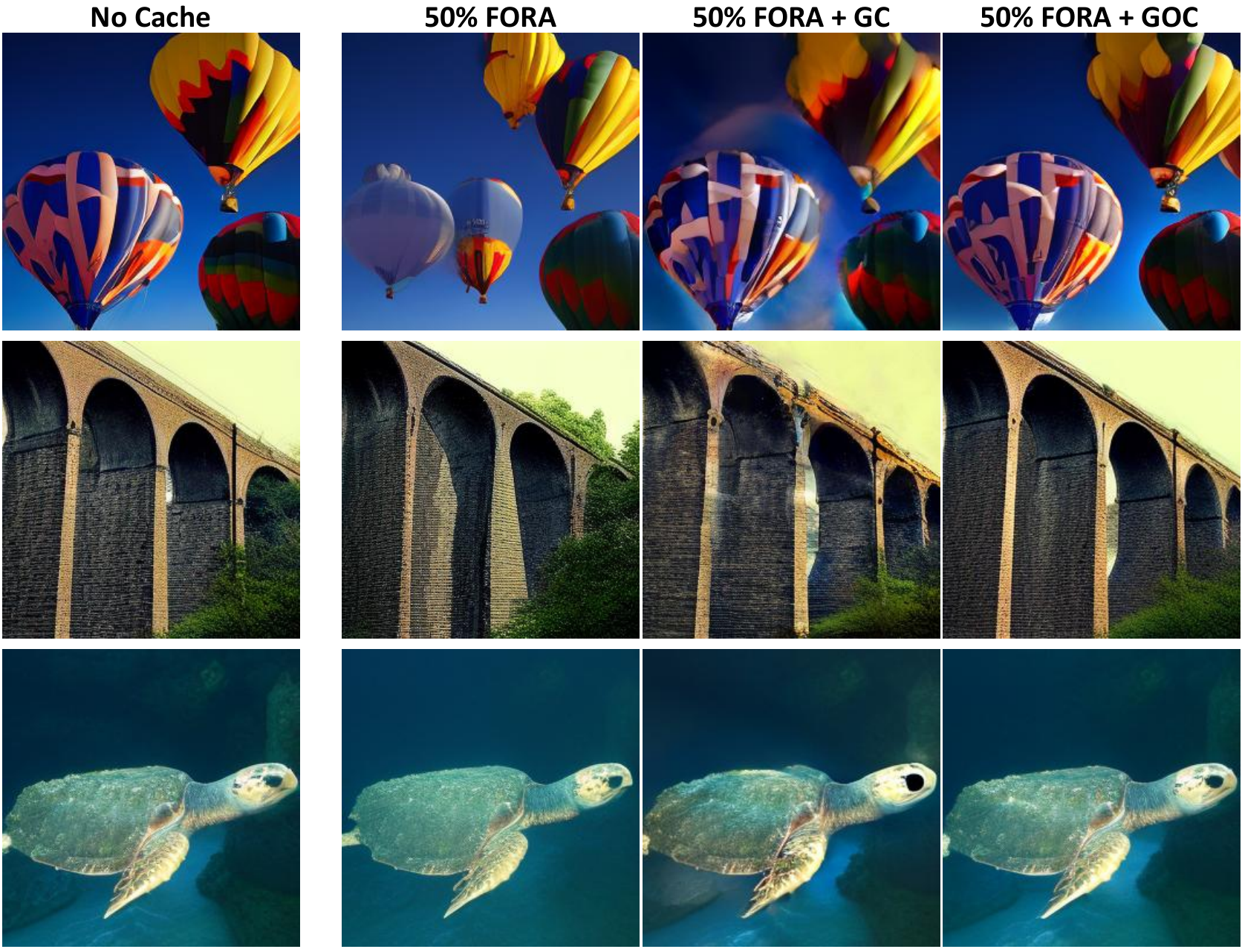}
   \caption{Comparison of generated images are compared under No Cache, 50\% FORA, 50\% FORA + GC, and 50\% FORA + GOC. Here, GC is used at each caching step.}
   \label{fig:RQ2}
\end{figure}

\begin{table*}[t]
    \centering
    \begin{tabular}{c|c|c|ccccc}
        \toprule
        \textbf{Cache Strategy} & \textbf{Caching level} & \textbf{$\eta$} & \textbf{IS↑} & \textbf{FID↓} & \textbf{sFID↓} & \textbf{Precision↑} & \textbf{Recall↑} \\ 
        \midrule
        \multirow{10}{*}{FORA\cite{selvaraju2024fora}+GOC} & \multirow{5}{*}{50\%}  & 1.4        & 204.760 & 4.432 & 5.623 & 0.752 & \textbf{0.599}  \\ 
        \cline{3-8} 
                        &                        & 1.3        & 213.479 & 3.940 & \textbf{4.862} & 0.767 & 0.588  \\ 
        \cline{3-8} 
                        &                        & 1.2        & 216.280 & \textbf{3.907} & 4.972 & 0.775 & 0.574  \\ 
        \cline{3-8} 
                        &                        & 1.1        & \textbf{216.941} & 4.015 & 5.262 & 0.776 & 0.571  \\ 
        \cline{3-8} 
                        &                        & 1          & 216.013 & 4.163 & 5.552 & \textbf{0.777} & 0.566  \\ 
        \cline{2-8}
                        & \multirow{5}{*}{25\%}  & 2.1        & 214.884 & 3.681 & 4.914 & 0.7693 & \textbf{0.593}  \\ 
        \cline{3-8} 
                        &                        & 2       & 218.581 & 3.526 & 4.635 & 0.776 & 0.587  \\ 
        \cline{3-8} 
                        &                        & 1.9        & 221.082 & \textbf{3.480} & \textbf{4.613} & 0.780 & 0.585  \\ 
        \cline{3-8} 
                        &                        & 1.8        & 222.210 & 3.498 & 4.694 & 0.783 & 0.580  \\ 
        \cline{3-8} 
                        &                        & 1.7          & \textbf{222.805} & 3.524 & 4.805 & \textbf{0.784} & 0.581  \\ 
        \midrule
        \multirow{5}{*}{L2C\cite{ma2024learning}+GOC} & \multirow{5}{*}{22\%}  & 1.2        & 235.369 & 3.205 & \textbf{4.629} & \textbf{0.802} & 0.558  \\ 
        \cline{3-8} 
                        & & 1.1        & 236.256 & 3.196 & 4.634 & 0.803 & \textbf{0.560}  \\ 
        \cline{3-8} 
                        & & 1        & \textbf{236.748} & \textbf{3.192} & 4.644 & 0.803 & 0.559  \\ 
        \cline{3-8} 
                        & & 0.9        & 236.653 & 3.202 & 4.652 & 0.805 & 0.558  \\ 
        \cline{3-8} 
                        &                        & 0.8          & 236.469 & 3.211 & 4.656 & 0.804 &  0.555 \\ 
        \bottomrule
    \end{tabular}
    \caption{Metrics of the generated images under FORA+GOC with 50\% and 25\% caching levels, as well as L2C+GOC with 22\% caching level, under different constraint of $\eta$.}
    \label{tab:RQ3}
\end{table*}

\begin{table*}[t]
    \centering
    \begin{tabular}{c|c|ccccc|cc}
\toprule
\textbf{Cache Strategy} & \textbf{Caching level} & \textbf{IS↑}               & \textbf{FID↓}               & \textbf{sFID↓}              & \textbf{Precision↑} & \textbf{Recall↑}  & \textbf{FLOPs(T)↓} & \textbf{Speed↑} \\ \midrule
No Cache\cite{peebles2023scalable}                & 0\%                    & 223.490                    & 3.484                       & 4.892                       & 0.788              & 0.571    & 11.868 & 1×       \\ \midrule
\multirow{2}{*}{FORA\cite{selvaraju2024fora}}     & 50\%                   & 190.046                    & 6.857                       & 8.757                       & 0.739               & 0.524     &5.934 &2.0000×       \\ \cline{2-9} 
                        & 25\%                   & 220.011                    & 3.870                       & 5.185                       & 0.783               & 0.569   &8.900 &1.3335×         \\ \midrule
\multirow{2}{*}{FORA+GOC} & 50\%                   & \textbf{216.280}           & \textbf{3.907}              & \textbf{4.972}              & \textbf{0.775}      & \textbf{0.574} &5.934 &1.9999×  \\ \cline{2-9} 
                        & 25\%                   & \textbf{222.805}           & \textbf{3.524}              & \textbf{4.805}              & \textbf{0.784}      & \textbf{0.581} &8.900 &1.3334×   \\ \midrule
L2C\cite{ma2024learning}                     & 22\%                   & 225.004                    & 3.539                       & 4.710                       & 0.788               & \textbf{0.563} &9.257 &1.2821×  \\ \midrule
L2C+GOC                 & 22\%                   & \textbf{236.748}           & \textbf{3.192}              & \textbf{4.644}              & \textbf{0.803}      & 0.559      &9.257 &1.2820×      \\ \bottomrule
\end{tabular}
    \caption{Metrics of the generated images before and after applying GOC to FORA or L2C under different caching levels.}
    \label{tab:RQ4-1}
\end{table*}

\subsection{Gradient Optimization Determination}
To verify the effectiveness and rationality of GOD, we design a two-part experiment. The first part demonstrates the improvements in metrics and images achieved by incorporating GOD, while the second part focuses on the selection of GOD parameters.

\noindent\textbf{Effect of GOD (RQ2)}.
To systematically evaluate caching optimization strategies, Table \ref{tab:RQ2} and Figure \ref{fig:RQ2} provide comparative analyses of four configurations: No Cache, FORA (Reduced Caching), FORA+GC (FORA with Gradient Caching), and FORA+GOD (GC-optimized FORA). Quantitative metrics in Table \ref{tab:RQ2} reveal that the 50\% FORA configuration significantly degrades all image quality parameters. Implementing GC across all caching stages substantially enhances performance, achieving optimal sFID (4.923) and Recall (0.593). The FORA+GOD configuration further refines this approach by selectively removing GC steps with adverse effects, attaining peak scores in IS (216.280), FID (3.907), and Precision (0.775).
By observing Figure \ref{fig:RQ2}, we can see that FORA, which skips half of the computational steps, causes the balloon to deform, the bridge arch to disappear, and insufficient details in the turtle. FORA+GC can restore the general shape of the image but introduces artifacts, making the balloon blurry, the bridge distorted, and the turtle's eyes unclear. FORA+GOD can avoid the artifacts of FORA+GC and generate images that closely resemble those of No Cache. While FORA+GC restores basic geometric structures, it introduces visible artifacts: blurred balloon surfaces, warped bridge geometries, and undefined ocular features in the turtle. In contrast, FORA+GOD achieves near-equivalent visual fidelity to the No Cache baseline, successfully resolving both the structural degradation of FORA and the artifact interference from FORA+GC. It demonstrates that GC implementation effectively compensates for FORA's quality degradation but introduces secondary artifacts. The GOD-enhanced configuration achieves optimal balance by retaining GC's corrective benefits while eliminating its error-prone steps, thereby maintaining structural coherence and metric superiority.

\begin{figure*}[t]
  \centering
\includegraphics[width=0.9\linewidth]{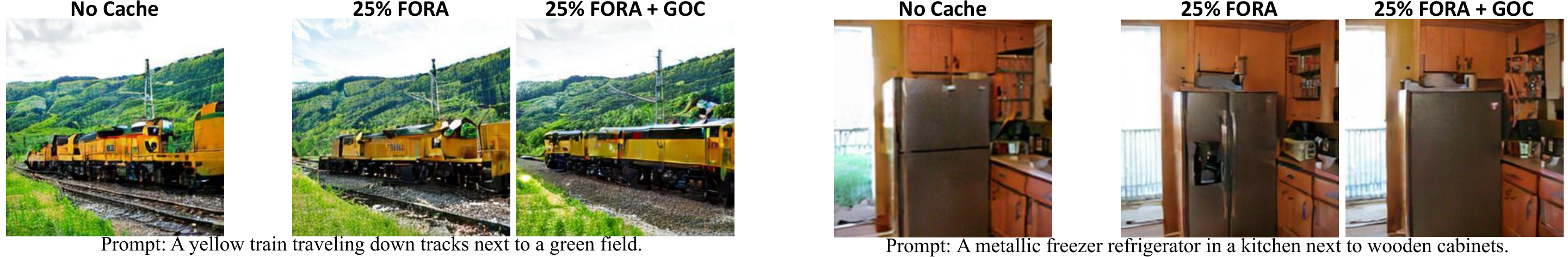}
   \caption{Comparison of images generated by the weight trained on the MS-COCO dataset under No Cache, 25\% FORA, and 25\% FORA + GOC conditions. We have attached the corresponding prompts below the images.}
   \label{fig:coco}
\end{figure*}

\noindent\textbf{Selection of hyperparameter (RQ3)}.
The parameter $\eta$ determines whether the gradient can maximally compensate for the error introduced by normal caching. As shown in Table \ref{tab:RQ3}, we select FORA+GOC with 50\% and 25\% of blocks cached, as well as L2C+GOC with 22\% of blocks cached. We record five values around the optimal $\eta$ for each case, with a step size of 0.1. We found that when $\eta = 1.2$, the 50\% FORA+GOC achieves the most balanced metrics. When $\eta = 1.9$, the 25\% FORA+GOC achieves the most balanced metrics. When $\eta = 1$, the 22\% L2C+GOC achieves the most balanced metrics. These different values of $\eta$ reveal a phenomenon: for the case of caching the first half of the odd steps (25\% FORA+GOC), a larger $\eta$ can more effectively reduce the caching error but also introduces a significant amount of gradient error. Since GOC is applied at the earlier steps, the model has sufficient steps to map the gradient error back to normal. For L2C+GOC, the sampling process includes knowledge obtained from training, which weakens the effect of GOC. Therefore, the value of $\eta$ is lower than in the other two cases.

\subsection{Comparison with Other Methods}
\textbf{Advantages and general applicability of GOC (RQ4)}.
As shown in Table \ref{tab:RQ4-1}, GOC can be applied to FORA and L2C with different caching levels and can improve the quality of generation. Specifically, in the case of 50\% FORA, GOC can significantly increase IS from 190.046 to 216.280, decrease FID from 6.857 to 3.907, and reduce sFID from 8.757 to 4.972. In the case of 25\% FORA, GOC can increase IS from 220.011 to 222.805, decrease FID from 3.870 to 3.524, and reduce sFID from 5.185 to 4.805. The improved metrics are very close to those of No Cache, and the sFID even exceeds it. In the case of 22\% L2C, GOC can significantly increase IS to 236.748 and decrease FID to 3.192. The improvement is greater than that of the similar proportion FORA, indicating that GOC and training-based caching methods can complement each other. In addition, we use FLOPs(T) to represent the computational load in the sampling process, and it can be seen that the computational cost brought by GOC is negligible.

\begin{table}[]
\centering
\begin{tabular}{c|c|c}
\toprule
\textbf{Cache Strategy} & \textbf{Caching level} & \textbf{FID↓}   \\ \midrule
No Cache\cite{chen2023pixart}              & 0\%                    & 21.964          \\ \midrule
FORA\cite{selvaraju2024fora}                      & \multirow{2}{*}{25\%}  & 21.984          \\ \cline{1-1} \cline{3-3} 
FORA+GOC                  &                        & \textbf{21.971} \\ \bottomrule
\end{tabular}
\caption{Metrics of the generated images generated by the weight trained on the MS-COCO dataset under No Cache, as well as FORA and FORA+GOC with 50\% caching level.}
\label{tab:RQ4-3}
\end{table}

To demonstrate that GOC is adaptable to both text-to-image and class-to-image tasks, we employ a Pixart model trained on the MS-COCO dataset and utilized the same strategy of caching 25\% of the blocks. We then examine the impact on images before and after incorporating GOC. As illustrated in Figure \ref{fig:coco}, GOC significantly compensates for the detail loss caused by 25\% FORA. For instance, GOC enhances the integrity of trains and refrigerators. In addition, Table \ref{tab:RQ4-3} shows that the introduction of GOC improves the FID score of FORA.

Finally, the original intention of caching is to reduce sampling time, so we need to pay attention to the increased computational load introduced by GOC, which may lead to longer sampling times. As shown in Table \ref{tab:RQ4-2}, the latency increases by approximately 2.0\%, 1.5\%, and 1.2\% for FORA+GOC (50\%), FORA+GOC (25\%), and L2C+GOC (22\%), respectively. Given the significant improvement in FID, these time costs are acceptable.

\begin{table}[t]
\centering
\begin{tabular}{c|c|c}
\toprule
\textbf{Cache Strategy}&\textbf{FID↓} & \textbf{latency(s)↓} \\ \midrule
FORA\cite{selvaraju2024fora} (50\%) & 6.857 & 1.789 ± 0.046 \\ \hline
FORA+GOC (50\%) & \textbf{3.907} & 1.824 ± 0.029 \\ \midrule
FORA (25\%) & 3.870 & 2.271 ± 0.032 \\ \hline
FORA+GOC (25\%) & \textbf{3.524} & 2.305 ± 0.016 \\ \midrule
L2C\cite{ma2024learning} (22\%) & 3.539 & 1.992 ± 0.011 \\ \hline
L2C+GOC (22\%) & \textbf{3.192} & 2.016 ± 0.004 \\ \bottomrule
\end{tabular}
\caption{Comparison of the average time taken to generate eight images under FORA and FORA+GOC with 50\% and 25\% caching levels, as well as L2C and L2C+GOC with 22\% caching level.}
\label{tab:RQ4-2}
\end{table}

\section{Conclusion}
In this paper, we propose Gradient-Optimized Cache (GOC), which leverages a queue to compute cached gradients during the sampling process and integrates these gradients, weighted appropriately, into the original cached features to reduce caching errors. 
Meanwhile, we utilize the model's statistical information to identify inflection points in the sampling process and combine step positions to avoid counter-gradient optimization, thereby further reducing gradient errors.
The proposed GOC can generate higher-quality samples with almost no additional computational cost. Through model comparisons and ablation studies, we achieve better performance under different datasets and various caching strengths for both FORA and L2C, thereby validating the effectiveness of our method.

Note that the implementation of GOC is still limited by fixed parameters and bound steps. Thus, we plan to investigate the development of a general embedding procedure for GOC in the future.
\section*{Acknowledgements}
This research is supported by the National Natural Science Foundation of China (No.U24B20180, No.62472393) and the advanced computing resources provided by the Supercomputing Center of the USTC.
{
    \small
    \bibliographystyle{ieeenat_fullname}
    \bibliography{main}

\begin{thebibliography}{44}
\providecommand{\natexlab}[1]{#1}
\providecommand{\url}[1]{\texttt{#1}}
\expandafter\ifx\csname urlstyle\endcsname\relax
  \providecommand{\doi}[1]{doi: #1}\else
  \providecommand{\doi}{doi: \begingroup \urlstyle{rm}\Url}\fi

\bibitem[Bhalgat et~al.(2020)Bhalgat, Lee, Nagel, Blankevoort, and Kwak]{bhalgat2020lsq}
Yash Bhalgat, Jaehoon Lee, Michael Nagel, Tijmen Blankevoort, and Nojun Kwak.
\newblock Lsq+: Improving low-bit quantization through learnable offsets and better initialization.
\newblock In \emph{Proceedings of the IEEE/CVF conference on computer vision and pattern recognition workshops}, pages 696--697, 2020.

\bibitem[Chadebec et~al.(2024)Chadebec, Tasar, Benaroche, et~al.]{chadebec2024flash}
Corentin Chadebec, Omer Tasar, Elie Benaroche, et~al.
\newblock Flash diffusion: Accelerating any conditional diffusion model for few steps image generation.
\newblock \emph{arXiv preprint arXiv:2406.02347}, 2024.

\bibitem[Chen et~al.(2023)Chen, Yu, Ge, Yao, Xie, Wu, Wang, Kwok, Luo, Lu, et~al.]{chen2023pixart}
Junsong Chen, Jincheng Yu, Chongjian Ge, Lewei Yao, Enze Xie, Yue Wu, Zhongdao Wang, James Kwok, Ping Luo, Huchuan Lu, et~al.
\newblock Pixart-$\alpha$: Fast training of diffusion transformer for photorealistic text-to-image synthesis.
\newblock \emph{arXiv preprint arXiv:2310.00426}, 2023.

\bibitem[Chen et~al.(2024)Chen, Shen, Ye, Cao, Tu, Bouganis, Zhao, and Chen]{chen2024delta}
Pengtao Chen, Mingzhu Shen, Peng Ye, Jianjian Cao, Chongjun Tu, Christos-Savvas Bouganis, Yiren Zhao, and Tao Chen.
\newblock ${\Delta-dit}$: A training-free acceleration method tailored for diffusion transformers.
\newblock \emph{arXiv preprint arXiv:2406.01125}, 2024.

\bibitem[Deng et~al.(2009)Deng, Dong, Socher, Li, Li, and Fei-Fei]{deng2009imagenet}
Jia Deng, Wei Dong, Richard Socher, Li-Jia Li, Kai Li, and Li Fei-Fei.
\newblock Imagenet: A large-scale hierarchical image database.
\newblock In \emph{2009 IEEE conference on computer vision and pattern recognition}, pages 248--255. Ieee, 2009.

\bibitem[Dong et~al.(2017)Dong, Chen, and Pan]{dong2017learning}
Xuanyi Dong, Shiyu Chen, and Sinno~Jialin Pan.
\newblock Learning to prune deep neural networks via layer-wise optimal brain surgeon.
\newblock In \emph{Advances in neural information processing systems}, 2017.

\bibitem[Gao et~al.(2023)Gao, Zhou, Cheng, and Yan]{gao2023masked}
Shanghua Gao, Pan Zhou, Ming-Ming Cheng, and Shuicheng Yan.
\newblock Masked diffusion transformer is a strong image synthesizer.
\newblock In \emph{Proceedings of the IEEE/CVF international conference on computer vision}, pages 23164--23173, 2023.

\bibitem[Guo et~al.(2019)Guo, Wang, Tian, and Wang]{guo2019dense}
Dan Guo, Shuo Wang, Qi Tian, and Meng Wang.
\newblock Dense temporal convolution network for sign language translation.
\newblock In \emph{IJCAI}, pages 744--750, 2019.

\bibitem[Guo et~al.(2022)Guo, Wang, Yu, McKenna, and Law]{guo2022adaln}
Yunhui Guo, Chaofeng Wang, Stella~X Yu, Frank McKenna, and Kincho~H Law.
\newblock Adaln: a vision transformer for multidomain learning and predisaster building information extraction from images.
\newblock \emph{Journal of Computing in Civil Engineering}, 36\penalty0 (5):\penalty0 04022024, 2022.

\bibitem[Ho et~al.(2020)Ho, Jain, and Abbeel]{ho2020denoising}
Jonathan Ho, Ajay Jain, and Pieter Abbeel.
\newblock Denoising diffusion probabilistic models.
\newblock \emph{Advances in neural information processing systems}, 33:\penalty0 6840--6851, 2020.

\bibitem[Lee et~al.(2019)Lee, Ajanthan, Gould, and Torr]{lee2019signal}
Namhoon Lee, Thalaiyasingam Ajanthan, Stephen Gould, and Philip~H Torr.
\newblock A signal propagation perspective for pruning neural networks at initialization.
\newblock \emph{arXiv preprint arXiv:1906.06307}, 2019.

\bibitem[Li et~al.(2023{\natexlab{a}})Li, Hu, Khan, Li, Yang, Wang, Cheng, and Yang]{li2023faster}
Senmao Li, Taihang Hu, Fahad~Shahbaz Khan, Linxuan Li, Shiqi Yang, Yaxing Wang, Ming-Ming Cheng, and Jian Yang.
\newblock Faster diffusion: Rethinking the role of unet encoder in diffusion models.
\newblock \emph{CoRR}, 2023{\natexlab{a}}.

\bibitem[Li et~al.(2023{\natexlab{b}})Li, Liu, Lian, Yang, Dong, Kang, et~al.]{li2023qdiffusion}
Xiangyu Li, Yujun Liu, Li Lian, Hui Yang, Zhen Dong, Dongdong Kang, et~al.
\newblock Q-diffusion: Quantizing diffusion models.
\newblock In \emph{Proceedings of the IEEE/CVF International Conference on Computer Vision}, pages 17535--17545, 2023{\natexlab{b}}.

\bibitem[Li et~al.(2021)Li, Gong, Tan, Yang, Hu, Zhang, et~al.]{li2021brecq}
Yixuan Li, Rui Gong, Xiaoyu Tan, Yibo Yang, Peng Hu, Qian Zhang, et~al.
\newblock Brecq: Pushing the limit of post-training quantization by block reconstruction.
\newblock \emph{arXiv preprint arXiv:2102.05426}, 2021.

\bibitem[Lin et~al.(2014)Lin, Maire, Belongie, Hays, Perona, Ramanan, Doll{\'a}r, and Zitnick]{lin2014microsoft}
Tsung-Yi Lin, Michael Maire, Serge Belongie, James Hays, Pietro Perona, Deva Ramanan, Piotr Doll{\'a}r, and C~Lawrence Zitnick.
\newblock Microsoft coco: Common objects in context.
\newblock In \emph{Computer vision--ECCV 2014: 13th European conference, zurich, Switzerland, September 6-12, 2014, proceedings, part v 13}, pages 740--755. Springer, 2014.

\bibitem[Liu et~al.(2024)Liu, Geddes, Guo, Jiang, and Nandwana]{liu2024smoothcache}
Joseph Liu, Joshua Geddes, Ziyu Guo, Haomiao Jiang, and Mahesh~Kumar Nandwana.
\newblock Smoothcache: A universal inference acceleration technique for diffusion transformers.
\newblock \emph{arXiv preprint arXiv:2411.10510}, 2024.

\bibitem[Liu et~al.(2021)Liu, Zhang, Kuang, Zhou, Xue, Wang, et~al.]{liu2021group}
Le Liu, Sheng Zhang, Zenghui Kuang, Aocheng Zhou, Jinhao Xue, Xiaogang Wang, et~al.
\newblock Group fisher pruning for practical network compression.
\newblock In \emph{International Conference on Machine Learning}, pages 7021--7032. PMLR, 2021.

\bibitem[Lu et~al.(2022)Lu, Zhou, Bao, Chen, Li, and Zhu]{lu2022dpm}
Chenlin Lu, Yihao Zhou, Fan Bao, Jianfei Chen, Chongxiao Li, and Jun Zhu.
\newblock Dpm-solver: A fast ode solver for diffusion probabilistic model sampling in around 10 steps.
\newblock In \emph{Advances in Neural Information Processing Systems}, pages 5775--5787, 2022.

\bibitem[Lu et~al.(2023)Lu, Wang, Zhang, Hao, and He]{lu2023semantic}
Jinda Lu, Shuo Wang, Xinyu Zhang, Yanbin Hao, and Xiangnan He.
\newblock Semantic-based selection, synthesis, and supervision for few-shot learning.
\newblock In \emph{Proceedings of the 31st ACM International Conference on Multimedia}, pages 3569--3578, 2023.

\bibitem[Lu et~al.(2025)Lu, Wu, Li, Jia, Wang, Zhang, Fang, Wang, and He]{lu2025damo}
Jinda Lu, Junkang Wu, Jinghan Li, Xiaojun Jia, Shuo Wang, YiFan Zhang, Junfeng Fang, Xiang Wang, and Xiangnan He.
\newblock Damo: Data-and model-aware alignment of multi-modal llms.
\newblock \emph{arXiv preprint arXiv:2502.01943}, 2025.

\bibitem[Ma et~al.(2024{\natexlab{a}})Ma, Fang, Mi, and Wang]{ma2024learning}
Xinyin Ma, Gongfan Fang, Michael~Bi Mi, and Xinchao Wang.
\newblock Learning-to-cache: Accelerating diffusion transformer via layer caching.
\newblock \emph{arXiv preprint arXiv:2406.01733}, 2024{\natexlab{a}}.

\bibitem[Ma et~al.(2024{\natexlab{b}})Ma, Fang, and Wang]{ma2024deepcache}
Xinyin Ma, Gongfan Fang, and Xinchao Wang.
\newblock Deepcache: Accelerating diffusion models for free.
\newblock In \emph{Proceedings of the IEEE/CVF Conference on Computer Vision and Pattern Recognition}, pages 15762--15772, 2024{\natexlab{b}}.

\bibitem[Meng et~al.(2022{\natexlab{a}})Meng, Bau, Andonian, and Belinkov]{meng2022locating}
Kevin Meng, David Bau, Alex Andonian, and Yonatan Belinkov.
\newblock Locating and editing factual associations in gpt.
\newblock \emph{Advances in Neural Information Processing Systems}, 35:\penalty0 17359--17372, 2022{\natexlab{a}}.

\bibitem[Meng et~al.(2022{\natexlab{b}})Meng, Sen~Sharma, Andonian, Belinkov, and Bau]{meng2022memit}
Kevin Meng, Arnab Sen~Sharma, Alex Andonian, Yonatan Belinkov, and David Bau.
\newblock Mass editing memory in a transformer.
\newblock \emph{arXiv preprint arXiv:2210.07229}, 2022{\natexlab{b}}.

\bibitem[Nagel et~al.(2020)Nagel, Amjad, Van~Baalen, Louizos, and Blankevoort]{nagel2020adaptive}
Michael Nagel, Rehan~A Amjad, Max Van~Baalen, Christos Louizos, and Tijmen Blankevoort.
\newblock Up or down? adaptive rounding for post-training quantization.
\newblock In \emph{International Conference on Machine Learning}, pages 7197--7206. PMLR, 2020.

\bibitem[Peebles and Xie(2023)]{peebles2023scalable}
William Peebles and Saining Xie.
\newblock Scalable diffusion models with transformers.
\newblock In \emph{Proceedings of the IEEE/CVF International Conference on Computer Vision}, pages 4195--4205, 2023.

\bibitem[Qiu et~al.(2025{\natexlab{a}})Qiu, Lu, and Wang]{qiu2025multimodal}
Junxiang Qiu, Jinda Lu, and Shuo Wang.
\newblock Multimodal generation with consistency transferring.
\newblock In \emph{Findings of the Association for Computational Linguistics: NAACL 2025}, pages 504--513, 2025{\natexlab{a}}.

\bibitem[Qiu et~al.(2025{\natexlab{b}})Qiu, Wang, Lu, Liu, Jiang, Zhu, and Hao]{qiu2025accelerating}
Junxiang Qiu, Shuo Wang, Jinda Lu, Lin Liu, Houcheng Jiang, Xingyu Zhu, and Yanbin Hao.
\newblock Accelerating diffusion transformer via error-optimized cache.
\newblock \emph{arXiv preprint arXiv:2501.19243}, 2025{\natexlab{b}}.

\bibitem[Salimans and Ho(2022)]{salimans2022progressive}
Tim Salimans and Jonathan Ho.
\newblock Progressive distillation for fast sampling of diffusion models.
\newblock \emph{arXiv preprint arXiv:2202.00512}, 2022.

\bibitem[Selvaraju et~al.(2024)Selvaraju, Ding, Chen, Zharkov, and Liang]{selvaraju2024fora}
Pratheba Selvaraju, Tianyu Ding, Tianyi Chen, Ilya Zharkov, and Luming Liang.
\newblock Fora: Fast-forward caching in diffusion transformer acceleration.
\newblock \emph{arXiv preprint arXiv:2407.01425}, 2024.

\bibitem[Shih et~al.(2024)Shih, Belkhale, Ermon, Sadigh, and Anari]{shih2024parallel}
Andrew Shih, Shashank Belkhale, Stefano Ermon, Dorsa Sadigh, and Nima Anari.
\newblock Parallel sampling of diffusion models.
\newblock In \emph{Advances in Neural Information Processing Systems}, 2024.

\bibitem[Song et~al.(2020)Song, Meng, and Ermon]{song2020denoising}
Jiaming Song, Chenlin Meng, and Stefano Ermon.
\newblock Denoising diffusion implicit models.
\newblock \emph{arXiv preprint arXiv:2010.02502}, 2020.

\bibitem[Song et~al.(2023)Song, Dhariwal, Chen, and Sutskever]{song2023consistency}
Yang Song, Prafulla Dhariwal, Mark Chen, and Ilya Sutskever.
\newblock Consistency models.
\newblock \emph{arXiv preprint arXiv:2303.01469}, 2023.

\bibitem[Wang et~al.(2018)Wang, Guo, Zhou, Zha, and Wang]{wang2018connectionist}
Shuo Wang, Dan Guo, Wen~gang Zhou, Zheng~jun Zha, and Meng Wang.
\newblock Connectionist temporal fusion for sign language translation.
\newblock In \emph{Proceedings of the 26th ACM international conference on Multimedia}, pages 1483--1491, 2018.

\bibitem[Wang et~al.(2025)Wang, Li, Mu, Hao, Liu, Wang, and He]{wang2025precise}
Yuan Wang, Ouxiang Li, Tingting Mu, Yanbin Hao, Kuien Liu, Xiang Wang, and Xiangnan He.
\newblock Precise, fast, and low-cost concept erasure in value space: Orthogonal complement matters.
\newblock In \emph{Proceedings of the Computer Vision and Pattern Recognition Conference}, pages 28759--28768, 2025.

\bibitem[Wimbauer et~al.(2024)Wimbauer, Wu, Schoenfeld, Dai, Hou, He, Sanakoyeu, Zhang, Tsai, Kohler, et~al.]{wimbauer2024cache}
Felix Wimbauer, Bichen Wu, Edgar Schoenfeld, Xiaoliang Dai, Ji Hou, Zijian He, Artsiom Sanakoyeu, Peizhao Zhang, Sam Tsai, Jonas Kohler, et~al.
\newblock Cache me if you can: Accelerating diffusion models through block caching.
\newblock In \emph{Proceedings of the IEEE/CVF Conference on Computer Vision and Pattern Recognition}, pages 6211--6220, 2024.

\bibitem[Wu et~al.(2024)Wu, Wang, Shang, Shah, and Yan]{wu2024ptq4dit}
Jiajun Wu, Haoyu Wang, Yilun Shang, Mubarak Shah, and Yan Yan.
\newblock Ptq4dit: Post-training quantization for diffusion transformers.
\newblock \emph{arXiv preprint arXiv:2405.16005}, 2024.

\bibitem[Yang et~al.(2024)Yang, Teng, Zheng, Ding, Huang, Xu, Yang, Hong, Zhang, Feng, et~al.]{yang2024cogvideox}
Zhuoyi Yang, Jiayan Teng, Wendi Zheng, Ming Ding, Shiyu Huang, Jiazheng Xu, Yuanming Yang, Wenyi Hong, Xiaohan Zhang, Guanyu Feng, et~al.
\newblock Cogvideox: Text-to-video diffusion models with an expert transformer.
\newblock \emph{arXiv preprint arXiv:2408.06072}, 2024.

\bibitem[Zhang et~al.(2024)Zhang, Xiao, Tang, et~al.]{zhang2024token}
En Zhang, Bing Xiao, Jie Tang, et~al.
\newblock Token pruning for caching better: 9 times acceleration on stable diffusion for free.
\newblock \emph{arXiv preprint arXiv:2501.00375}, 2024.

\bibitem[Zheng et~al.(2023)Zheng, Nie, Vahdat, Azizzadenesheli, and Anandkumar]{zheng2023fast}
Hengrui Zheng, Weijia Nie, Arash Vahdat, Kamyar Azizzadenesheli, and Anima Anandkumar.
\newblock Fast sampling of diffusion models via operator learning.
\newblock In \emph{International conference on machine learning}, pages 42390--42402. PMLR, 2023.

\bibitem[Zhu et~al.(2025)Zhu, Tang, Liu, et~al.]{zhu2025dip}
Heng Zhu, Di Tang, Jie Liu, et~al.
\newblock Dip-go: A diffusion pruner via few-step gradient optimization.
\newblock \emph{Advances in Neural Information Processing Systems}, 37:\penalty0 92581--92604, 2025.

\bibitem[Zhu et~al.(2024{\natexlab{a}})Zhu, Wang, Lu, Hao, Liu, and He]{zhu2024boosting}
Xingyu Zhu, Shuo Wang, Jinda Lu, Yanbin Hao, Haifeng Liu, and Xiangnan He.
\newblock Boosting few-shot learning via attentive feature regularization.
\newblock In \emph{Proceedings of the AAAI Conference on Artificial Intelligence}, pages 7793--7801, 2024{\natexlab{a}}.

\bibitem[Zhu et~al.(2024{\natexlab{b}})Zhu, Zhu, Tan, Wang, Hao, and Zhang]{zhu2024enhancing}
Xingyu Zhu, Beier Zhu, Yi Tan, Shuo Wang, Yanbin Hao, and Hanwang Zhang.
\newblock Enhancing zero-shot vision models by label-free prompt distribution learning and bias correcting.
\newblock \emph{Advances in Neural Information Processing Systems}, 37:\penalty0 2001--2025, 2024{\natexlab{b}}.

\bibitem[Zhu et~al.(2024{\natexlab{c}})Zhu, Zhu, Tan, Wang, Hao, and Zhang]{zhu2024selective}
Xingyu Zhu, Beier Zhu, Yi Tan, Shuo Wang, Yanbin Hao, and Hanwang Zhang.
\newblock Selective vision-language subspace projection for few-shot clip.
\newblock In \emph{Proceedings of the 32nd ACM International Conference on Multimedia}, pages 3848--3857, 2024{\natexlab{c}}.

\end{thebibliography}
}

\end{document}